# Metaphors' journeys across time and genre:
# tracking the evolution of literary metaphors with temporal embeddings

Veronica Mangiaterra[1], Chiara Barattieri di San Pietro[1], Paolo Canal[1], Valentina Bambini[1]

[1]Laboratory of Neurolinguistics and Experimental Pragmatics (NEPLab), Department of Humanities and Life Sciences, University School for Advanced Studies IUSS, Pavia, Italy

**Abstract**
Metaphors are a distinctive feature of literary language, yet they remain less studied experimentally than everyday metaphors. Moreover, previous psycholinguistic and computational approaches overlooked the temporal dimension, although many literary metaphors were coined centuries apart from contemporary readers. This study innovatively applies tools from diachronic distributional semantics to assess whether the processing costs of literary metaphors varied over time and genre. Specifically, we trained word embeddings on literary and nonliterary Italian corpora from the 19th and 21st centuries, for a total of 124 million tokens, and modeled changes in the semantic similarity between topics and vehicles of 515 19th-century literary metaphors, taking this measure as a proxy of metaphor processing demands. Overall, semantic similarity, and hence metaphor processing demands, remained stable over time. However, genre played a key role: metaphors appeared more difficult (i.e., lower topic-vehicle similarity) in modern literary contexts than in 19th-century literature, but easier (i.e., higher topic-vehicle similarity) in today's nonliterary language (e.g., the Web) than in 19th-century nonliterary texts. This pattern was further shaped by semantic features of metaphors' individual terms, such as vector coherence and semantic neighborhood density. Collectively, these findings align with broader linguistic changes in Italian, such as the stylistic simplification of modern literature, which may have increased metaphor processing demands, and the high creativity of the Web's language, which seems to render metaphor more accessible.

## 1. Introduction

Research on metaphors, both from the perspective of cognitive linguistics (Lakoff & Johnson, 1980) and pragmatics (Sperber & Wilson, 2012), has largely contributed to depicting metaphors as a pervasive phenomenon in language and cognition, rather than just a tool for poets. Similarly, empirical research as pursued within Experimental Pragmatics, psycholinguistics, and neurolinguistics has focused on everyday manifestations of metaphors. Nonetheless, metaphor remains a stylistic hallmark of poetic and narrative language (G. J. Steen et al., 2010). Understanding the mechanisms through which we comprehend metaphors in literature captures, therefore, a relevant aspect of metaphor processing. To date, however, experimental research on literary metaphors remains a niche.

Theoretically, literary metaphors hold a special status compared to everyday metaphors. According to Relevance Theory, while everyday metaphors point toward a straightforward interpretation (e.g., 'That lawyer is a shark', meaning 'The lawyer is aggressive'), literary metaphors are held to generate a wide range of weak implicatures (e.g., the metaphor 'His ink is pale' by Gustave Flaubert, meaning that 'His writing lacks contrast, may fade, will not last'[1]). Processing this wide range of assumptions is supposed to require an additional cognitive effort that generates the so-called *poetic effect* of literary metaphors (Pilkington, 2000). On the quantitative side, corpus-linguistics studies found that metaphors are well represented in fiction, where they constitute 11% of the words (G. J. Steen et al., 2010), and that metaphors in literary texts are, more frequently than in other genres, non-lexicalized, i.e., with a novel association between topic and vehicle (Goatly, 1997).

From the experimental point of view, the few available studies on literary metaphors led to heterogeneous results. Katz et al. (1988) collected ratings from human participants for literary and everyday metaphors, finding no difference along ten psycholinguistic dimensions and suggesting that they do not differ in nature. Conversely, other studies reported that literary metaphors are less familiar, more open-ended, and more difficult to understand than journalistic metaphors (Semino & Steen, 2008; G. Steen, 1994). In addition, Bambini et al. (2014) found that ratings are modulated by context, and specifically that when literary metaphors are presented together with their context, people tend to consider them more meaningful and less concrete, less difficult, and less familiar, compared to when presented without context. Consistently, electrophysiological evidence showed that literary metaphors trigger a sustained brain response, possibly linked to the manipulation of multiple meanings (Bambini et al., 2019), supporting the theoretical claims that literary metaphors evoke multiple weak implicatures and require a greater cognitive effort compared to nonliterary metaphors.

A less considered factor in the study of literary metaphors is the temporal dimension: literary metaphors used in empirical studies were created by authors who lived well before the participants who took part in the experiments investigating the processing of these expressions. Whether this temporal gap between the time of production of literary metaphors and the time of their processing contributes to the cognitive effort involved in interpreting metaphors, and whether readers today need an additional effort compared to readers contemporary to the time of metaphor production, has never been investigated. A preliminary answer may come from the replication of the Katz et al. (1988) study by Campbell & Raney (2016). They recollected ratings for the same metaphors used in Katz et al. twenty-five years later and found that judgments were consistent over time. However, twenty-five years is a limited time span, while the time in which metaphors were created and contemporary readers might be centuries apart.

The issue of the temporal gap becomes particularly relevant when we consider that both language and society have deeply changed over the last two centuries, potentially modifying the conceptual and contextual factors underlying metaphor interpretation. Focusing on Italian, studies on the history of language noted that in the 19th century, written language was still an élite product. Despite some innovative thrusts, Italian was characterized by a style close to the courtly and illustrious tradition both in literature (Beccaria, 1993; Serianni, 1993) and in journalistic and nonfiction language (Marazzini, 2002; Masini, 1977). In the 21st century, instead, language has evolved, particularly in literary contexts, toward a strong shift to oral and informal styles (Coletti, 2022). Moreover, in the 21st century, much of the exposure to written language occurs on the Web, as highlighted, for instance, by recent reports on news consumption that indicate that web-based sources are used by 68% of the Italian sample, compared to 12% for print media (Newman et al., 2025). The language of the web, or *netspeak* (Crystal, 2006), has its own specificities: it is based on informal speech, combining terms from different registers and neologisms (Cerruti & Onesti, 2013), and allows users to experiment with new forms of language creatively (Fiorentino, 2018; Goddard, 2015).

---
[1] Example from Pilkington (2000).

### 1.1. *A distributional approach to metaphor evolution*

To understand how literary metaphors were perceived at the time of their creation, distributional semantics techniques can be of help, as they allow us to construct representations of meaning mimicking the semantic networks of readers of different epochs and to compare the status of the metaphors in the two representations. Two key features of these techniques make them suitable for answering our research question: i) their ability to approximate human representation of meaning and ii) the possibility to build models starting from historical corpora and therefore represent meaning in specific time points. Distributional semantics is based on the theoretical intuition, the so-called *distributional hypothesis* (Harris, 1954), that states that words occurring in similar contexts have similar meanings. Implementing this hypothesis, Vector Space Models (VSMs) can be trained on large text corpora to learn co-occurrences of words. In VSMs, words are represented as vectors, whose coordinates are derived from their co-occurrences in the corpora. Words whose vectors are closer in the VSM tend to be closer in meaning. In these models, the semantic similarity between two words can be operationalized as the cosine of the angle between the vector of word$_1$ ($v_{w1}$) and the vector of word$_2$ ($v_{w2}$), as defined by the following formula:

$$Cosine\ Similarity = \cos(v_{w1}, v_{w2}) = \frac{v_{w1} \cdot v_{w2}}{\|v_{w1}\| \cdot \|v_{w2}\|}$$

#### 1.1.1. *Applications of distributional semantics to psycholinguistic and metaphor research*

Since their inception, distributional semantic approaches have been closely tied to psychological and psycholinguistic research, given that VSMs can model various dimensions of human semantic knowledge, capturing how word meaning is acquired, processed, and stored in the brain (Bhatia et al., 2019; Günther et al., 2016; Jones et al., 2015). Distributional semantics tools can be a valid application to quantify also aspects of figurative language processing (Reid & Katz, 2018). In these applications, a widely studied measure is the semantic similarity between the two terms of a metaphor. Well before the diffusion of computational tools, the relation between topic, namely the subject of the metaphor (e.g., "lawyer" in "That lawyer is a shark"), and vehicle, namely the term used metaphorically (e.g., "shark"), was recognized as crucial to metaphor comprehension and appreciation, with rating-based semantic similarity being linked to a number of dimensions, such as metaphoricity, ease of interpretation, and goodness of nonliterary and literary metaphors (Katz et al., 1985; Marschark et al., 1983). The computational operationalization of semantic similarity via word embeddings has been largely applied to metaphor research (Bolognesi & Aina, 2019; Brglez & Vintar, 2025; Utsumi, 2011). Applications ranged from automatic metaphor identification (based on defining a cosine similarity threshold below which expressions are considered metaphorical, see Mao et al., 2018; Shutova, 2015; Su et al., 2017) to modeling metaphor processing costs (where cosine similarity is considered an approximation of human judgments). For instance, McGregor et al. (2019) found that contextual embeddings, i.e., dynamic representations of words based on surrounding contexts, can efficiently model human ratings of *metaphoricity*, *meaningfulness*, and *familiarity,* and Winter & Strik-Lievers (2023) showed that semantic similarity between the topic and the vehicle of synesthetic metaphors mirrors their degree of *metaphoricity* and *creativity* come from the *Figurative Archive* (Bressler et al., 2026), a recent resource collecting approximately 1,000 metaphors together with human ratings, where significant correlations between semantic distance and both familiarity and metaphoricity were reported. In particular, metaphors with less semantically similar topics and vehicles are considered less familiar and more difficult to process.

From word embeddings, other measures relevant for language and metaphor processing can be derived. Among those, *Semantic Neighborhood Density* (SND), i.e., the number of words that are similar to the target one in terms of meaning, was shown to play a role in language processing, being able to predict performances on psycholinguistic tasks, such as lexical decision and word naming (Buchanan et al., 2001). In metaphor research, metaphors composed of low-SND words were considered more comprehensible than metaphors with high-SND words (Al-Azary & Buchanan, 2017). Moreover, the SND of the vehicle allows to discriminate between different types of metaphors: literary metaphors have vehicles with higher SND compared to nonliterary metaphors (Reid et al., 2023).

### 1.1.2. *Applications of distributional semantics to semantic change*

Semantic similarity has found applications in historical linguistics and semantic change research. The *distributional hypothesis* can indeed be adapted to the diachronic perspective: changes in a word's co-occurrences reflect changes in its meaning (Hilpert, 2008). Hence, by looking at how word co-occurrences change through time, it is possible to track the evolution of meaning in time. Operationally, to create time-characterized word embeddings, i.e., embeddings representing the word's meaning at each specific time point, it is necessary to train VSMs on corpora of different epochs. However, given the stochastic nature of VSMs, every time a training process is initialized, different spatial coordinates are used, making it impossible to compare word vectors across spaces. For the VSMs to be comparable, they need to have the same spatial coordinates, i.e., to be aligned. Different procedures (Gulordava & Baroni, 2011; Kulkarni et al., 2015) have been proposed to train aligned VSMs, among which the *Temporally aligned Word Embeddings with a Compass* (TWEC) model (Di Carlo et al., 2019).

By employing aligned time-locked semantic representations of a word, it is possible to track its semantic shift over time by computing the cosine similarity between the embedding at time$_1$ ($v_{wt1}$) and at time$_2$ ($v_{wt2}$), a measure called *vector coherence* (Cassani et al., 2021; Hamilton et al., 2016). As a result, a word whose meaning has shifted over time would exhibit a lower semantic similarity between $v_{wt1}$ and $v_{wt2}$ (i.e., lower vector coherence) compared to a word whose meaning has remained consistent, which would exhibit a higher semantic similarity between $v_{wt1}$ and $v_{wt2}$ (higher vector coherence). To date, most studies in diachronic distributional semantics have focused on identifying semantic changes of single words (Cain & Ryskin, 2025; Charlesworth et al., 2022; Garg et al., 2018; Hamilton et al., 2016; Xu & Kemp, 2015), by examining variations in vector coherence across epochs. Only recently these approaches have been applied to investigate how the relation between different words changes over time (Jenkins et al., 2025), a perspective that is crucial for tracing the evolution of complex expressions, such as metaphors.

### 1.2. *The present study*

In this study, we aimed to explore whether Italian literary metaphors are associated with different processing demands for today's readers compared to 19[th]-century readers, contemporary to the time of metaphors' original creation. To do so, we innovatively extended diachronic VSMs from word-level to the case of multi-word expressions, namely metaphors. Specifically, we trained VSMs on historical corpora from the 19[th] century and on contemporary corpora, mimicking the linguistic input available to present and past readers. Given that (i) the evolution of Italian is closely intertwined with textual genres, and (ii) different interpretative attitudes and processing modes may be activated depending on the literariness of a text (G. Steen, 1989), we incorporated the genre dimension by training separate models for literary and nonliterary corpora within each epoch. Then, we compared the semantic similarity between topics and vehicles of a set of literary metaphors in each diachronic VSM, taking this measure as a proxy of the evolution of metaphors' processing demands. Moreover, we examined whether metaphors' evolution is further shaped by lexical-semantic features of individual topics and vehicles, such as their stability over time (vector coherence), their semantic neighborhood density, and their frequency.

We expected that the temporal dimension, when assessed over a sufficiently long span of time, would significantly influence the processing costs of metaphors. In particular, we hypothesized that metaphors would yield higher elaboration demands in present-day readers than in readers of the past, who shared the conceptual and contextual environment in which the metaphors were originally created. Also, given that in the past both literary and nonliterary texts were intended for educated readerships and exhibited similar stylistic features (Aprile, 2014), we expected no difference across genres in the past. In the 21$^{st}$ century, instead, literary language and web-based language constitute distinct varieties of Italian, which led us to expect differences between the two genres in the processing demands of metaphors.

## 2. Methods
### 2.2. Metaphor dataset

A set of 19$^{th}$-century Italian literary texts was retrieved from Project Gutenberg (https://www.gutenberg.org/), an initiative started in the 1970s, intending to collect digital versions of books that have never been copyrighted or whose copyright has lapsed. From the selected texts, "A di B" (Eng. "A of B") strings were extracted. Using the spaCy package (Honnibal & Montani, 2017), we performed a Part-Of-Speech (POS) tagging, and all the "NOUN of NOUN" strings were selected. Even though metaphors come in many shapes and forms, we chose to focus on the "NOUN of NOUN" structure to avoid any possible confounding factors due to differences in the number and type of their constituting elements. Following (Hanks, 2006), a set of keywords from semantic classes that are considered productive sources of metaphors, such as natural events and locations (river, storm, rain) and emotions (anger, joy), was then employed to filter the resulting list, yielding a set of N = 400 metaphors. The dataset was further enriched by adding an existing collection of "A of B"[2] metaphors (Bambini et al., 2014). The final dataset included a total of N = 515 metaphors in the form of "A of B" (e.g., "Capelli di fiamma", Eng. "Hair of flame") and is fully available in the *Literary Metaphors* module of the *Figurative Archive* (Bressler et al., 2026). While metaphors can come in different syntactic constructions, we chose the genitive one because it clearly displayed the two terms (topic and vehicle) of the metaphors. The order of the terms can vary, with some metaphors displaying a topic-vehicle (TV) order, with the first term being the topic and the second one being the vehicle, e.g., "Capelli di fiamma" (Eng. "Hair of flame"), and others displaying a vehicle-topic (VT) order, with the first term being the vehicle and the second one being the topic, e.g., "Grumo di nuvole" (Eng. "Clump of clouds"). Table 1 reports four examples of the metaphors included in the dataset, together with metadata regarding author, source, year of publication, topic, vehicle, and their order.

**Table 1.** Examples of literary metaphors included in the final dataset of the study.

| Metaphor | Author | Source/Book | Year | Topic | Vehicle | Order |
|---|---|---|---|---|---|---|
| Cielo di perla (Eng. sky of pearl) | Giovanni Pascoli | Myricae | 1891 | Cielo (Eng. Sky) | Perla (Eng. Pearl) | TV |
| Grumo di nuvole (Eng. clump of clouds) | Federico De Roberto | L'Illusione | 1891 | Nuvole (Eng. Clouds) | Grumo (Eng. Clump) | VT |
| Capelli di fiamma (Eng. hair of flame) | Sibilla Aleramo | Il Passaggio | 1919 | Capelli (Eng. Hair) | Fiamma (Eng. Flame) | TV |
| Nebbia di malinconia (Eng. Fog of melancholy) | Federico De Roberto | Documenti Umani | 1888 | Malinconia (Eng. Melancholy) | Nebbia (Eng. Fog) | VT |

**Note.** T = Topic, V = Vehicle.

---

[2] Thirty-seven of these metaphors displayed a definite artile in the propositional phrase, i.e., "A of *the* B" (e.g., "Abbraccio del sonno", Eng. "Embrace of the sleep").

### 2.1. Training sets

To examine the diachronic evolution of metaphors, considering also the possible effect of textual genre, we collected a set of Italian corpora, for a total of 124 million tokens, composed of literary and nonliterary texts written in the 19th and the 21st century. Table 2 reports a summary of the characteristics of the collected corpora.

Table 2. The composition of the literary and nonliterary sections of 19th-century and 21st-century corpora.

| Corpus | Size | Description | Genre |
| --- | --- | --- | --- |
| **21st Century** | | | |
| Itwac | 42 M | Web-crawled corpus (Baroni et al., 2009). | Nonliterary |
| Contemporary Literature | 20 M | Literary texts published between 2000 and 2020, used in accordance with the "fair use" principle of copyright law. | Literary |
| **19th century** | | | |
| Gutenberg Nonliterary (GNL) | 10 M | Nonliterary texts downloaded from Project Gutenberg on topics such as botany, agriculture, and science. | Nonliterary |
| Lessico dell'Italiano Scritto (LIS) | 6 M | Diachronic corpus of written Italian texts from 1850 to 1940 (Accademia della Crusca, 2013). | Nonliterary |
| ChronicItaly (CI) | 16 M | Corpus of Italian immigrant newspapers published in the United States between 1898 and 1920 (Viola, 2021). | Nonliterary |
| Gutenberg Literary | 30 M | Literary texts downloaded from Project Gutenberg, both prose and poetry. | Literary |

**Note.** Size is in millions of tokens.

From the corpora, four training sets were built: 19th-Century Literary training set, 19th-Century Nonliterary training set, 21st-Century Literary training set, 21st-Century Nonliterary training set. The training sets included from 20 to 42 million tokens (19th century literary: 30 M; 19th century nonliterary: 32 M; 21st century literary 20 M; 21st century nonliterary: 42 M), in line with the mean size of large corpora employed by Di Carlo et al. (2019). The internal diversity of these datasets was designed to provide a proxy for the linguistic input accessible to the readers in each respective epoch. In the 19th-century training set, we included prose and poetry texts for the literary section and texts from technical manuals, newspapers, and diaries for the nonliterary section. In the 21st-century training set, we included prose and poetry for the literary section and a collection of texts taken from the web, such as newspaper sites, blogs, and educational sites, for the nonliterary section.

### 2.2. Training Aligned Spaces

Employing the *Temporally aligned Word Embeddings with a Compass* (TWEC) model (Di Carlo et al., 2019), we trained four sets of word embeddings on the four training sets previously outlined. The word embeddings provide a semantic representation of words in the Italian language that differs by epoch (19th and 21st centuries) and genre (Literary and Nonliterary).

The TWEC model exploits the dual representation of words derived from a word2vec model (Mikolov et al., 2013) based on a *Continuous Bag of Words* (CBOW) architecture, a feed-forward neural network trained to predict a target word given its context, relying on the theoretical assumption that most words do not change over time, and that words with a shifted meaning will appear in the context of words that did not change. This theoretical assumption is reflected in the creation of an atemporal VSM, called *compass*, based on which the spatial coordinates of all the other temporal VSMs are then subsequently defined. In other words, a *compass* model is first trained on the entire corpus, providing the semantic representation of words independently of time. After that, the *compass*'s context matrix is used to initialize the training of a time- and genre-specific target matrix on each corpus, which allows us to extract the temporal word embeddings.

Operationally, we trained the model on the whole training set (resulting from the combination of all training sets), and we extracted the resulting two atemporal matrices, an input-weight *Context* matrix (C) and an output-weight *Target* matrix (U). The target matrix U was used as "a compass", i.e., it served as a reference to initialize all the other VSMs along the same coordinate system. We hence trained a set of context matrices $C_i$ on each slice of the training set (namely, 19$^{th}$-century Literary training set, 19$^{th}$-Century Nonliterary training set, 21$^{st}$-Century Literary training set, 21$^{st}$-Century Nonliterary training set). As a result, the context embeddings can differ according to the co-occurrence frequencies that are specific to a given temporal period. Each of the four resulting sets of word embeddings provides a representation of word meaning in the 19$^{th}$ century, in the 21$^{st}$ century, and in the literary and nonliterary sections of the training sets.

### *2.3. Measures of Diachronic Change*

To semantically characterize the metaphors across time, we computed four measures of interest using the obtained sets of word embeddings, one at the metaphor level (Cosine Similarity between topic and vehicle - CS) and three at the word level, for each term of the metaphor (Semantic Neighborhood Density – SND, Vector Coherence – VC, and Frequency – Freq, see Table 3). Each measure was computed, for each metaphor, in all time and genre slices, to obtain a semantic characterization of the metaphor and its terms in each epoch and each textual genre. A graphical representation of the approach, including corpora collection, VSM training, and features computation, is depicted in Figure 1.

**Table 3.** Measures of metaphor change, with corresponding formula and interpretation.

| Level | Measures | Formula | Description | Interpretation |
|---|---|---|---|---|
| **Metaphor** | Cosine Similarity between topic and vehicle (CS) | $CS_{(tv)} = \cos(v_t^{tn}, v_v^{tn})$ | Cosine Similarity between the vector of the metaphor's Topic ($v_t$) and the vector of the metaphor's Vehicle ($v_v$) in all time and genre slices ($t_n$). | CS can be considered as a proxy of the processing costs of the metaphor; from a diachronic perspective, it can be used to describe the temporal trajectories of the metaphors' demands. |
| **Word** | Semantic Neighborhood Density (SND) | $SND_t = \frac{\sum_{i=1}^{n=500} \cos(v_t^{tn}, v_i^{tn})}{n}$ $SND_v = \frac{\sum_{i=1}^{n=500} \cos(v_v^{tn}, v_i^{tn})}{n}$ | Semantic Neighborhood Density of the Topic ($v_t$) and the Vehicle ($v_v$) in all the slices ($t_n$) was computed as the average of the cosine similarities between the word and its $n$ closest neighbors. | SND refers to the average proximity of a word vector to its closest neighbors as computed using a language model. It provides a measure of the word's position in VSMs relative to its nearest neighbors. A word with many close neighbors is considered semantically denser than a word with fewer close neighbors. Metaphors with denser vehicles are considered less comprehensible (Al-Azary & Buchanan, 2017). |
| | Vector Coherence (VC) | $VC_t = \cos(v_t^{t1}, v_t^{t2})$ $VC_v = \cos(v_v^{t1}, v_v^{t2})$ | Vector Coherence of the Topic and the Vehicle was obtained by computing the cosine similarity between a word vector in the 19th century ($v_w^{t1}$) and its vector in the 21st century ($v_w^{t2}$). | VC is a measure of the stability of word meaning over time. A word with high VC has maintained a stable meaning because the word vector at $t_1$ is very close to the word vector at $t_2$. A word with low VC has changed meaning because the word vector at $t_1$ is quite far from the word vector at $t_2$. |
| | Frequency (Freq) | $Freq_t = \log\left(\frac{\text{number of occurrences topic}}{\text{total number of token}}\right)$ $Freq_v = \log\left(\frac{\text{number of occurrences vehicle}}{\text{total number of token}}\right)$ | Logarithmic Relative Frequency of the Topic ($Freq_t$) and the Vehicle ($Freq_v$) in all slices. | Word Frequency is a key measure both in psycholinguistics and in diachronic semantic shift research (Baumann et al., 2023). Regarding word processing, higher-frequency words are elaborated faster than lower-frequency ones (Brysbaert et al., 2018). In diachrony, frequent words show the tendency to change more slowly (Hamilton et al., 2016). |

**Note.** The value of *n* was set to 500, as in previous computational approaches to metaphors (Kintsch, 2000).

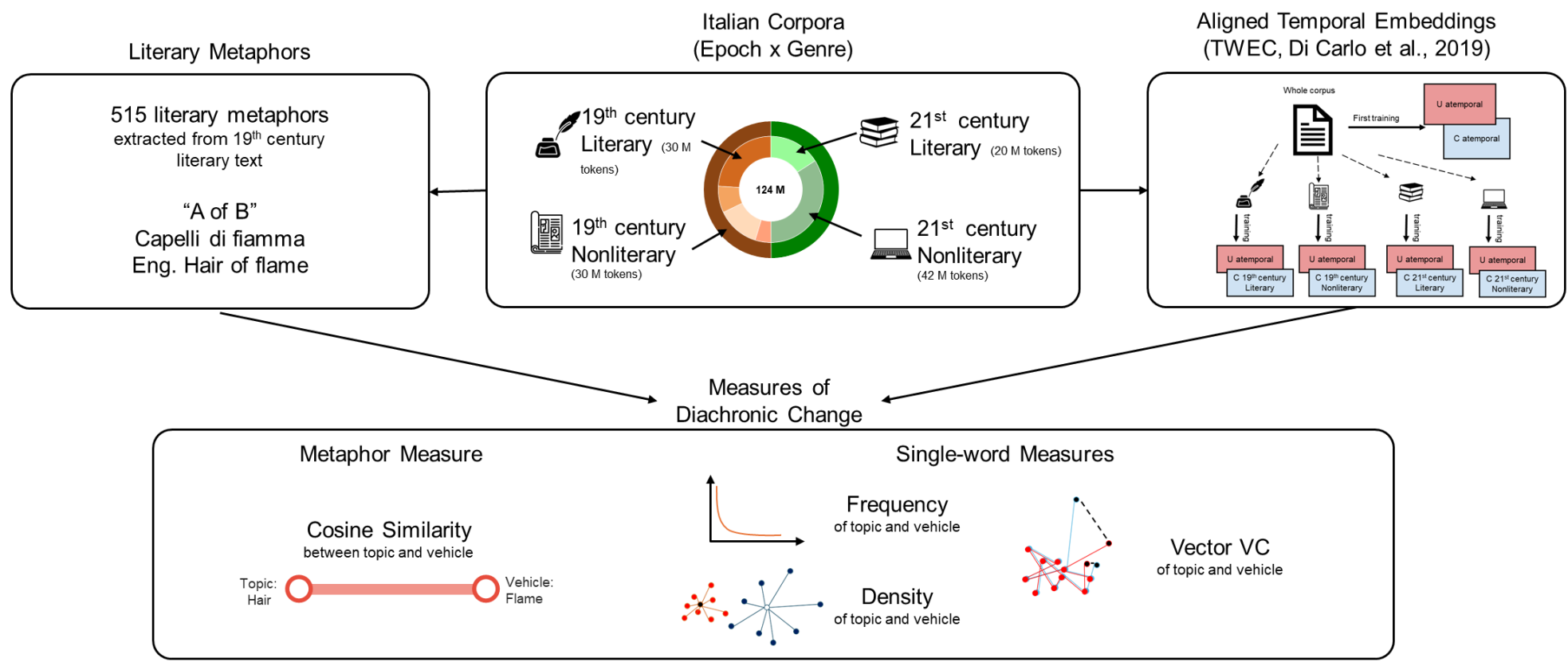

**Figure 1. Workflow for the Diachronic Analysis of Italian Literary Metaphors.** The process begins with the collection of Italian corpora stratified by epoch (19th vs. 21st century) and genre (literary vs. nonliterary). From the 19th-century literary corpus, 515 metaphors (e.g., Hair of flame) were extracted. Then, the corpora were used to train aligned temporal embeddings (using the TWEC methodology). Diachronic change is then quantified using a metaphor-level measure (Cosine Similarity) and single-word measures, including Frequency, Semantic Neighborhood Density (SND), and Vector Coherence (VC).

*2.4. Statistical analyses*

To understand the relationship between word- and metaphor-level measures, we computed a set of Pearson correlations between CS, SND, VC, and Freq, corrected for multiple comparisons.

To test the hypothesis that CS varied through time and across different genres, we fitted a Linear Mixed-Effects Model (LMM – Pinheiro & Bates, 2000) using the *lme4* package (Bates et al., 2015), considering CS as a continuous dependent variable, and genre and epoch as interacting categorical predictors. Then, to explore whether the evolution of metaphors is further shaped by different word-level variables, we fitted a series of LMMs using the *lme4* package again, considering CS as a continuous dependent variable, genre and epoch as interacting categorical predictors, and word-level variables as continuous predictors. Specifically, to define the final statistical model, we incrementally added the word-level variables (i.e., first topic and vehicle SND, Freq, and VC separately, then pairs of topic and vehicle features, and finally all the variables), and we tested if the addition of variables contributed to explaining the data variance by comparing the models' goodness-of-fit in terms of Akaike Information Criterion (AIC), Bayesian Information Criterion (BIC), and Log-likelihood (Bozdogan, 1987; Neath & Cavanaugh, 2012). All models included a random intercept to account for the variability of individual metaphors. All analyses were performed in R (R Core Team, 2025).

## 3. Results
### *3.1. Descriptive statistics*

Table 4 reports descriptive statistics of the measures of interest as computed in the four corpora and sets of embeddings. The distribution of each variable is displayed in Figure 2. While measures like CS, SND, and Freq maintain relatively normal distributions with consistent peaks across both centuries, VC exhibits a significant departure from normality. Specifically, in the literary genre, both topics and vehicles demonstrate high VC (approaching 1.0), suggesting that words in literary texts did not report drastic changes in meaning. Conversely, the words in the nonliterary genres show a great semantic shift.

**Table 4**. Descriptive statistics of word-level and metaphor-level measures reported as Mean (SD).

|  | 19th Lit | 21st Lit | 19th NonLit | 21st NonLit |
|---|---|---|---|---|
| **CS** | 0.31 (0.21) | 0.27 (0.21) | 0.30 (0.22) | 0.33 (0.20) |
| **SND Topic** | 0.69 (0.05) | 0.69 (0.05) | 0.68 (0.05) | 0.71 (0.05) |
| **SND Vehicle** | 0.71 (0.05) | 0.69 (0.05) | 0.70 (0.05) | 0.72 (0.05) |
| **VC Topic** | 0.85 (0.1) | | -0.08 (0.17) | |
| **VC Vehicle** | 0.84 (0.12) | | -0.12 (0.18) | |
| **Freq Topic** | -9.62 (1.45) | -9.90 (1.67) | -10.16 (1.47) | -10.69 (1.58) |
| **Freq Vehicle** | -10.53 (1.45) | -10.68 (1.49) | -11.09 (1.56) | -11.57 (1.51) |

**Note:** CS = Cosine Similarity, computed between the Topic and the Vehicle of the metaphor; SND = Semantic Neighborhood Density, computed as average cosine similarity between the word and its 500 closest neighbors; VC = Vector Coherence, computed as cosine similarity between the word at $t_1$ and the word at $t_2$; Freq = Frequency, computed as the logarithm of the relative frequency of the word.

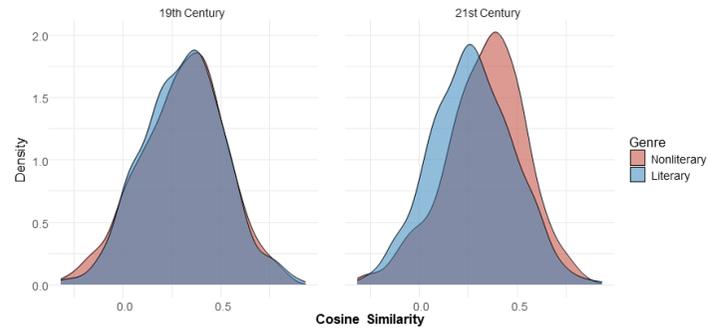
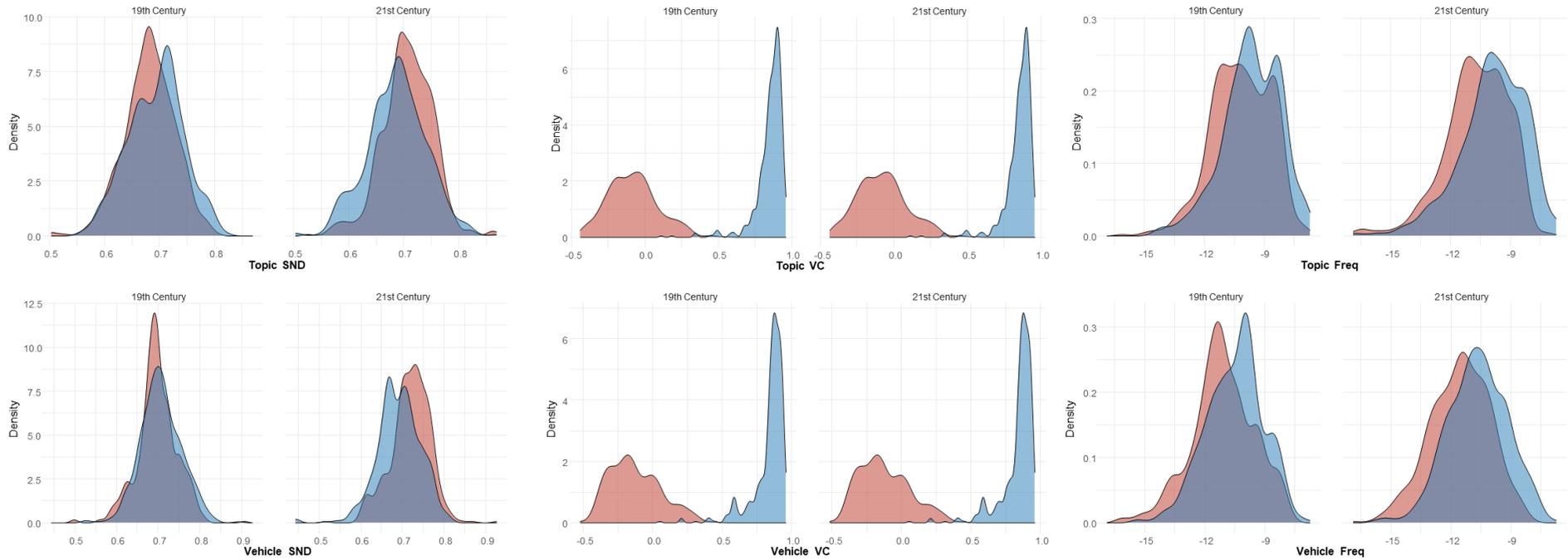

**Figure 2. Density plots of metaphor and single-word variables.** The plots show the density distribution of the metaphor feature (cosine similarity) and single-word features (semantic neighborhood density, vector coherence, and frequency) for both topic and vehicle across time and genre. Note: SND = Semantic Neighborhood Density, VC = Vector Coherence, Freq = Frequency.

### *3.2. Correlation analysis*

Results of the correlation analyses between single-word and metaphor-level measures are reported in Figure 3.

CS, SND, and Freq showed overall strong positive correlations across genre and epoch, as highlighted by the black triangles on the diagonal. VC, however, reported low correlations, confirming the different pattern of semantic change between literary and nonliterary genres.

Regarding word-level variables, SND and Freq of both topics and vehicles were negatively correlated in all slices (see the blue squares), indicating that less frequent words tend to be more semantically dense across time and genre. A positive correlation emerged between Freq and VC (see the light blue rectangles), especially in literary corpora, indicating that more frequent words have a more stable meaning. Moreover, SND positively correlated with VC (see the purple rectangle), suggesting that denser words have a more stable meaning.

Regarding the relations between word-level variables and CS (see green rectangles), we found that the latter was positively correlated with topic VC in literary corpora, suggesting that when the meaning of the topics tends to change over time, metaphors are characterized by more semantically distant terms. Moreover, in nonliterary corpora, we found a positive correlation between both topic and vehicle SND and CS, indicating that when metaphors are constituted by high-density terms, they tend to have a greater topic-vehicle similarity.

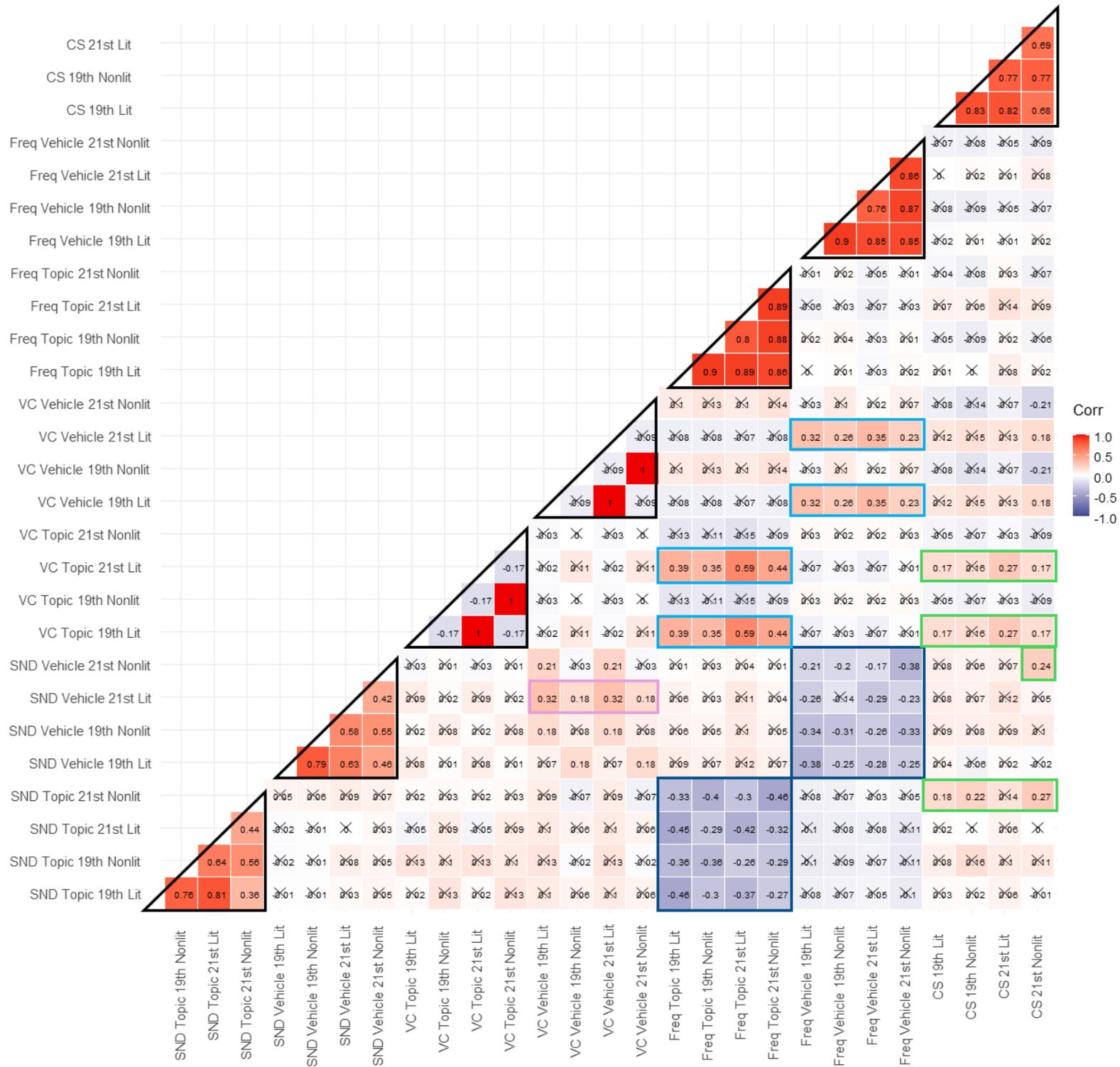

**Figure 3. Correlations between word-level measures and metaphor-level measures.** Positive correlations are displayed in red and negative correlations in blue. The color intensity is proportional to the correlation coefficients. Non-significant correlations are marked with a cross. All corrections are corrected for multiple comparisons.

### *3.3. Linear Mixed-Effects Models*

The simple LMM including epoch and genre as interacting predictors revealed an effect of Genre ($\beta$ = -0.03, $t$ = -6.85, $p < 0.001$), with metaphors reporting a lower semantic similarity in Literary corpora compared to Nonliterary ones, further qualified by its interaction with Epoch ($\beta$ = -0.07, $t$ = -7.82, $p < 0.001$). No main effect of Epoch was reported ($p$ = 0.35). A diverging trend emerged: metaphors' terms become increasingly distant in literary texts going from the 19th century to the 21st century, while in nonliterary texts, metaphors' terms showed an increasing semantic similarity (Table 6, Figure 4).

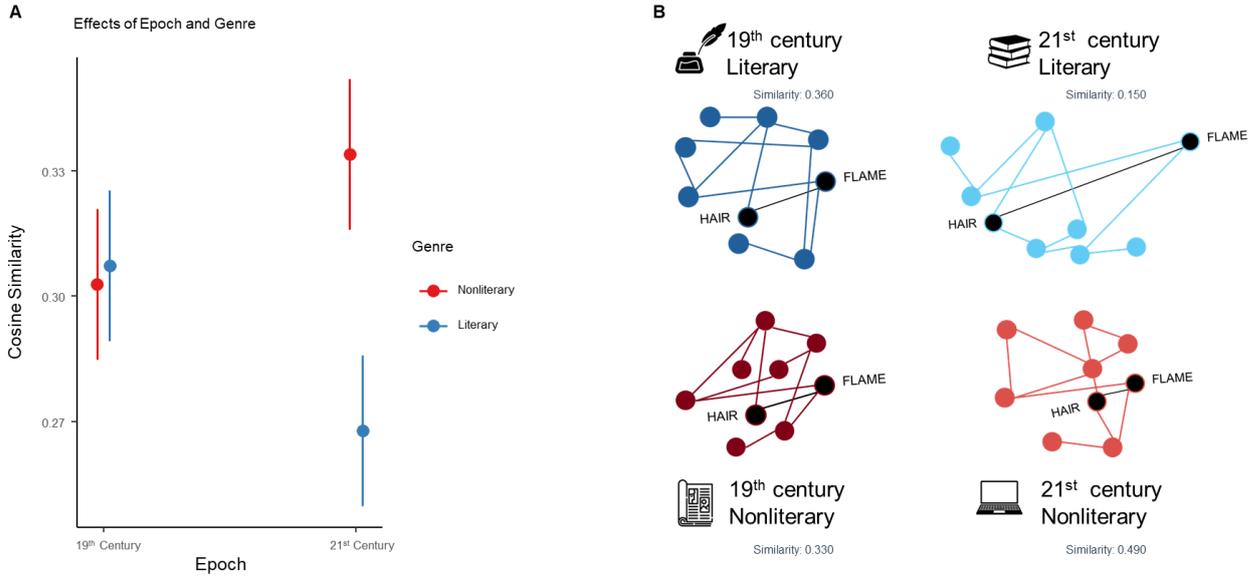

**Figure 4. Effects of epoch and genre on cosine similarity between the topics and vehicles of metaphors.** Panel A shows the effects of epoch and genre as emerged from the LMM. Panel B shows a graphical representation of how the relationship between topic and vehicle of a representative metaphor ("Capelli di fiamma", Eng. "Hair of flame") changed.

Moving to the investigation of how the diachronic evolution of metaphors is further shaped by lexical-semantic features of topic and vehicle, the comparison of AIC, BIC, and Log-likelihood showed that the best fit to the data is obtained by adding both topic and vehicle SND and VC to the simple model (see Table 5).

**Table 5.** Summary statistics of LMMs presenting AIC, BIC, and Log-Likelihood.

| Model | df | AIC | BIC | LogLik |
| --- | --- | --- | --- | --- |
| Epoch * Genre | 6 | -2,185.8 | -2,152.1 | 1,098.9 |
| Epoch * Genre * (SND topic + SND vehicle) | 14 | -2,321.5 | -2,242.8 | 1,174.8 |
| Epoch * Genre * (Freq topic + Freq vehicle) | 14 | -2,212.3 | -2,133.5 | 1,120.1 |
| Epoch * Genre * (VC topic + VC vehicle) | 14 | -2,231.3 | -2,152.6 | 1,129.7 |
| **Epoch * Genre * (VC topic + VC vehicle + SND topic + SND vehicle)** | **22** | **-2,357.1** | **-2,233.4** | **1,200.5** |
| Epoch * Genre * (VC topic + VC vehicle + Freq topic + Freq vehicle) | 22 | -2,245.3 | -2,121.6 | 1,144.7 |
| Epoch * Genre * (SND topic + SND vehicle + Freq topic + Freq vehicle) | 22 | -2,328.0 | -2,204.3 | 1,186.0 |
| Epoch * Genre * (VC topic + VC vehicle + SND topic + SND vehicle + Freq topic + Freq vehicle) | 30 | -2,349.8 | -2,181.2 | 1,204.9 |

**Note:** AIC = Akaike's information criterion; BIC = Bayesian Information Criterion; Freq = word Frequency; LogLik = Log-Likelihood; SND = Semantic Neighborhood Density; VC = Vector Coherence.

The LMM showed a significant three-way interaction between topic VC, Genre, and Epoch ($\beta = 0.225$, $t = 3.15$, $p = 0.002$), suggesting that the coherence of the topic influenced metaphor CS in contemporary literary texts but not in the other slices (Table 6, Figure 5a). Moreover, the model showed a significant interaction between vehicle SND, Genre, and Epoch ($\beta = -0.602$, $t = -3.25$, $p = 0.001$), in addition to the main effect of vehicle SND ($\beta = 0.58$, $t = 7.90$, $p < 0.001$). While the main effect indicated that higher SND generally predicts higher CS, this relationship became stronger specifically in 21$^{st}$ century nonliterary texts (Table 6, Figure 5b).

**Table 6.** Outputs of the base LMM model and the best LMM model with Single-Word Features on CS.

| Predictors | Base Model | | | Model with Single-Word Predictors | | |
|---|---|---|---|---|---|---|
| | *Estimates* | *Statistic* | *p* | *Estimates* | *Statistic* | *p* |
| Epoch | -0.00 | -0.93 | 0.353 | -0.28 | -2.99 | **0.003** |
| Genre | -0.03 | -6.85 | **<0.001** | 0.15 | 1.55 | 0.121 |
| Epoch*Genre | -0.07 | -7.82 | **<0.001** | 0.11 | 0.59 | 0.552 |
| SND Topic | | | | 0.57 | 7.63 | **<0.001** |
| SND Vehicle | | | | 0.58 | 7.90 | **<0.001** |
| VC Topic | | | | 0.04 | 1.92 | 0.056 |
| VC Vehicle | | | | -0.04 | -2.12 | **0.034** |
| Epoch*SND Topic | | | | 0.02 | 0.24 | 0.811 |
| Epoch* SND Vehicle | | | | 0.23 | 2.48 | **0.013** |
| Epoch * VC Topic | | | | 0.12 | 3.30 | **0.001** |
| Epoch * VC Vehicle | | | | -0.04 | -1.44 | 0.150 |
| Genre * SND Topic | | | | -0.25 | -2.63 | **0.008** |
| Genre * SND Vehicle | | | | -0.16 | -1.72 | 0.086 |
| Genre * VC Topic | | | | 0.23 | 4.35 | **<0.001** |
| Genre * VC Vehicle | | | | 0.08 | 1.76 | 0.079 |
| Epoch * Genre * SND Topic | | | | 0.18 | 0.99 | 0.321 |
| Epoch * Genre * SND Vehicle | | | | -0.60 | -3.25 | **0.001** |
| Epoch * Genre * VC Topic | | | | 0.23 | 3.15 | **0.002** |
| Epoch * Genre * VC Vehicle | | | | 0.02 | 0.33 | 0.740 |

**Note:** SND = Semantic Neighborhood Density; VC = Vector Coherence.

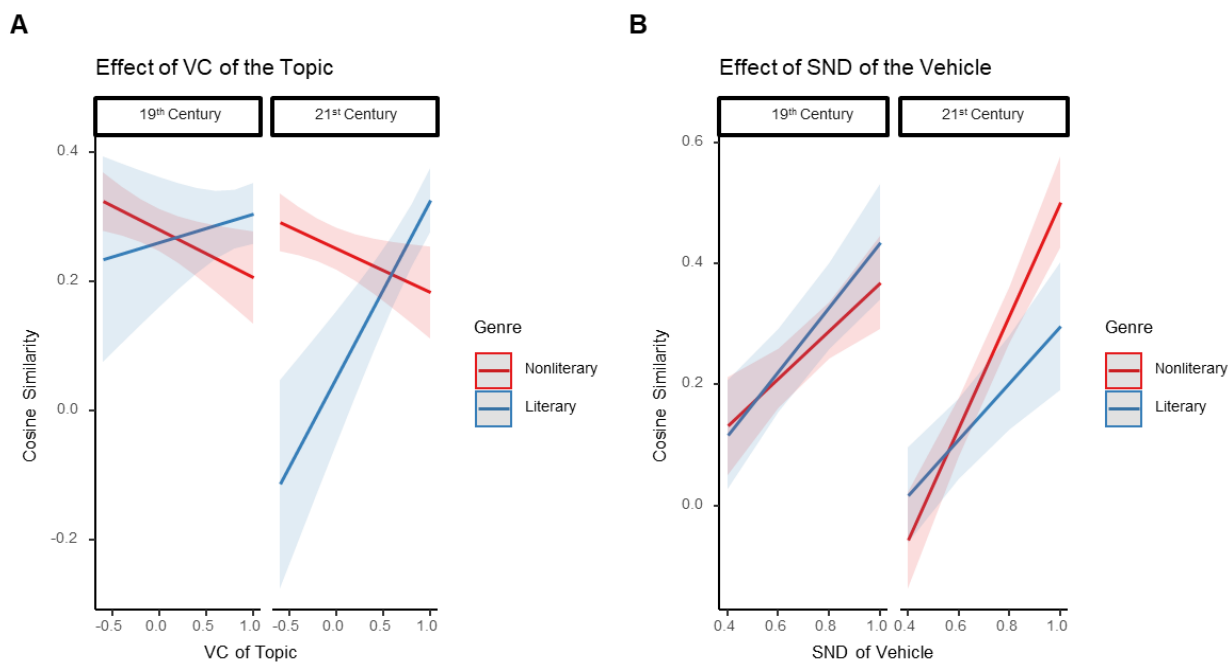

**Figure 5. Significant effects of lexical-semantic features of topic and vehicle.** Panel A illustrates the effect of Vector Coherence (VC) of the topic on Cosine Similarity (CS) across Epoch and Genre. Panel B shows the effect of Semantic Neighborhood Density (SND) of the vehicle on CS across Epoch and Genre.

## 4. Discussion

In the present study, we investigated whether the processing demands of Italian literary metaphors changed over time and as a function of textual genres and lexical-semantic features of the topic and vehicle. To do so, we collected 19th-century and 21st-century corpora (composed of literary and nonliterary texts), taken to reflect the linguistic input of today's and past readers. Then, we used the corpora to train different temporal vector space models, obtaining representations of word meanings at the two time points. Then, we examined how the semantic similarity between topics and vehicles of a set of 515 Italian metaphors, extracted from literary texts of the 19th century, varied. Crucially, we took semantic similarity as a proxy of metaphor processing difficulty, thereby associating metaphors with semantically closer terms to lower processing demands. Importantly, we did not limit the analysis to semantic similarity across epochs and genres, but we also included three single-word variables relevant in historical and metaphor research. Specifically, we considered i) a measure of the stability of word meaning over time (*vector coherence*; Cassani et al., 2021; Hamilton et al., 2016b; Rodda et al., 2017); ii) *word frequency*, as it affects both semantic shift (Englhardt et al., 2020; Feltgen et al., 2017; Sagi et al., 2009) and metaphor processing (Littlemore et al., 2018); iii) Semantic Neighborhood Density (SND), which has been shown to impact metaphor processing, with high density hindering metaphor comprehensibility (Al-Azary & Buchanan, 2017; Reid et al., 2023). Given the significant stylistic change and simplification of Italian over the past two centuries (Coletti, 2022), we expected that today's readers may experience classical literary metaphor differently from the readers of the 19th century. In particular, we assumed that today's readers do not share the cultural background of the authors of classical literary metaphors, and, as such, we expected that metaphors could entail greater processing costs for contemporary readers.

Our analysis partially confirmed our predictions. Overall, the semantic relationship between topics and vehicles of metaphors did not change from the 19th century to today, suggesting that the processing of metaphors has not become more costly as we moved away from the time these metaphors were originally written. However, a different representation of the metaphors across time emerges depending on the genre (literary or nonliterary). Indeed, the significant interaction between Epoch and Genre indicates that the semantic relationship between topic and vehicle followed a genre-dependent diverging trajectory over time. Starting from an equivalent semantic representation in 19th-century literary and nonliterary VSMs, metaphors' terms in the 21st-century became increasingly distant (less similar) in literary texts, while they showed an increasing semantic similarity in nonliterary texts. This interaction suggests that the effort required to process these metaphors has actually decreased in modern nonliterary contexts; conversely, in the literary domain, these metaphors have become more semantically distant over time.

This result can be linked to the nature of the corpora considered as literary and nonliterary texts. The modern nonliterary corpus consisted of web-crawled texts (Baroni et al., 2009), a linguistic variety that differs substantially from literature (Pistolesi, 2014). Scholars have noted the language of the Web is characterized by greater variability, brevity, fragmentation, fluidity, and loose use of language with an open, hybrid, and changeable nature (Santini, 2007), while modern Italian literature is marked by a loss of literariness and a shift toward orality, highlighted by lexicon and discursive signals specific to a standardized spoken language (Dardano, 2014). In the modern nonliterary corpus, the vectors of topics and vehicles tend to be closer, indicating more entrenched associations and easier connections between distant concepts. In the literary modern corpus, by contrast, the vectors tend to be more distant, suggesting that novel associations are less frequent and therefore more striking. We can conclude that in contemporary nonliterary language, novel metaphorical associations required reduced processing demand due to increased frequency of novel associations between words, while in literary language, they remain marked and, hence, cognitively demanding.

It is noteworthy that, while in the 21st century the status of metaphors is quite distinct across genres, in line with the marked stylistic differences between literary and nonliterary texts reported for modern Italian (Aprile, 2014), this distinction was not present in the 19th-century texts. Indeed, the interaction effect suggests that in the 19th century, metaphors were processed similarly regardless of the textual genre. This can be linked to the shared stylistic features of 19th-century literary and nonliterary texts. Texts from both genres were cultural products designed for the educated portion of the population and therefore characterized by high-register language, with echoes of the classical tradition (Aprile, 2014). Although elements of stylistic innovation can be found starting from the 19th century, language essays, manuals, and newspaper texts, which compose our nonliterary training sets, continue the expressive heritage of earlier prose and poetry (Masini, 1994).

Interestingly, in a proof-of-concept study applying the same methodology to English literary metaphors (Mangiaterra et al., 2024), we found a different pattern of results. English metaphors were associated with higher processing costs in nonliterary text compared to literary ones, irrespective of the epoch, confirming the stable nature of the English language – and its stylistic remarks, such as metaphors – in the last two centuries. These cross-linguistic differences suggest that the patterns emerging from this kind of analysis are language-specific and that metaphor evolution follows the broader stylistic trajectories of the language in which they are embedded.

### *4.1. Single-word features and their impact on metaphor evolution*

The claim that metaphor evolution is embedded in the broader temporal changes of the language at stake is further strengthened when we consider the role of single-word variables. First, it is important to note that single-word variables showed known patterns of relationships, confirming the validity of our vector space models. In particular, vector coherence was associated with word frequency, consistent with the findings that frequent words change less over time (Hamilton et al., 2016b), and with semantic neighborhood density, consistent with the findings that words change more in the sparse portions of the semantic network (Ryskina et al., 2020). We also found that word frequency was negatively correlated with SND (consistent with Buchanan et al., 2001), suggesting that highly frequent words tend to occur in a variety of contexts and develop more loosely related relationships with their neighbors (less dense meaning), while low-frequency words tend to develop a higher level of specificity and tighter relationship with their neighbors (more dense meaning, see Rambelli & Bolognesi, 2024).

Moving to the analysis of the contribution of lexical-semantic features of topics and vehicles in shaping the diachronic evolution of metaphors, both the correlation analysis and the linear mixed models highlighted the centrality of the vehicle semantic neighborhood density and the topic vector coherence. Semantic neighborhood density of the vehicle had an overall effect, which appeared stronger in 21st century nonliterary text, as denser vehicles were associated with metaphors with semantically closer terms. As suggested by Al-Azary & Buchanan (2017), when a word has many close semantic neighbors, it lacks the flexibility to acquire new metaphorical associations. Instead, a vehicle with low semantic neighborhood density, with its looser semantic association, may evoke the wide array of weak implicatures characteristic of literary metaphors, which resulted in greater appreciation (Reid et al., 2023), but also higher processing costs, as suggested by our results. In the 21st century nonliterary text, high SND pulls the terms together into the entrenched, easily processed associations reflected in our high similarity results, such as in the highly dense vehicle "flour" in the metaphor "sky of flour", as compared to the sparser vehicle "inebriation" in the metaphor "inebriation of light".

Moreover, we found that in 21st-century literary texts, where metaphors are associated with higher processing costs, these demands seem to be driven by the vector coherence of the topic. In deriving the meaning of a metaphor, the function of the topic is to help promote the salient features of the vehicle necessary to reach the intended interpretation. Topics whose meanings have changed greatly over time have probably acquired a sparser set of semantic relationships and are considered less concrete (Azarbonyad et al., 2017) and acquired later (Cassani et al., 2021). All these features contribute to making the shifted meaning of these topics less accessible and less prone to providing the straightforward contextual constraints necessary to guide the reader through the process of meaning derivation of complex literary metaphors. An example of this process can be provided by the metaphor "vento di lode" (Eng. "wind of praise"). The topic "lode" (Eng. "praise") shifted greatly between literary texts of the 19th-century (where the meaning was in the semantic domain of "kindness", "gratitude", "appreciation") and 21st-century (where the meaning is in the semantic domain of "report card", "degree", "graduated", given the prominence of the use in the expression "con lode" – Eng. "cum laude"), and therefore represented a less clear guide for the selection of the appropriate meaning of the metaphor, increasing its processing demands.

Overall, it emerged that the topic should have a stable meaning to guide the derivation of metaphor interpretation over time and reduce processing costs, and, at the same time, the vehicle should have a flexible meaning to allow the metaphor to emerge.

5. **Conclusions**

This study examined the evolution of processing costs of literary metaphors in terms of semantic similarity between topics and vehicles, expanding the diachronic application of word embeddings to the analysis of complex expressions such as metaphors. Our results showed that, while overall the processing demands of metaphors did not change from the past to today, they vary in relation to different textual genres, following the broader patterns of stylistic evolution in the Italian language. We can therefore argue that since literary and nonliterary language were very similar in the 19th century, readers did not process metaphors with different levels of effort depending on genre. Today's readers, however, have to make a greater effort to process metaphors in current literary texts, which have a plainer and simplified language, while they can more easily activate the connection between distant concepts in the creative nonliterary language of the Web. Possibly, this distinction may also hint at the effect of different linguistic exposure and its impact of processing in contemporary speakers with different backgrounds.

Methodologically, this work confirms the possibility of applying temporal embeddings to examine the evolution of multi-word expressions (Jenkins et al., 2025), extending their scope to figurative expressions. Regarding metaphor processing, our results empirically highlight the need to account for the textual context in which metaphors are embedded, which could determine different processes of elaboration (G. Steen, 1989).

The point of optimal distance between topics and vehicles, which "should be sufficiently distant to emphasize differences but sufficiently close to maintain similarities" (Katz, 1989), seems to be the result of a complex balance, modulated by the nature of the words composing the metaphors but also sensitive to the broader stylistic features of the language. Ultimately, this study demonstrates that what makes a metaphor difficult to process is not an inherent property, but a dynamic process influenced by diachronic genre-based shifts and by the structure of single-word semantic networks.

**Acknowledgment**

This work received support from the European Research Council under the EU's Horizon Europe programme, ERC Consolidator Grant "PROcessing MEtaphors: Neurochronometry, Acquisition and Decay, PROMENADE" (GA: 101045733). The content of this article is the sole responsibility of the authors. The European Commission or its services cannot be held responsible for any use that may be made of the information it contains.

**Code and data availability statement**

Temporal vector space models, scripts, and metaphor datasets used in the study are available at the anonymized link https://zenodo.org/records/18523748.